\documentclass[letterpaper]{article}
\usepackage{aaai}
\usepackage{times}
\usepackage{helvet}
\usepackage{courier}

\usepackage{algorithmic, amssymb, amsmath}   
\usepackage{algorithm}
\usepackage{multirow}
\usepackage{graphicx}
\usepackage[T5,T1]{fontenc}
\usepackage{multirow}
\usepackage{array}

\begin{document}

\title{Automatically Creating a Large Number of New Bilingual Dictionaries}
\author{Khang Nhut Lam \and Feras Al Tarouti \and Jugal Kalita\\
Computer Science Department \\
University of Colorado, USA \\
\{klam2, faltarou, jkalita\}@uccs.edu\\
}
\maketitle
\begin{abstract}
\begin{quote}
This paper proposes approaches to automatically create a large number of new bilingual dictionaries for low-resource languages, especially resource-poor and endangered languages, from a single input bilingual dictionary. Our algorithms produce translations of words in a source language to plentiful target languages using available Wordnets and a machine translator (MT). Since our approaches rely on just one input dictionary, available Wordnets and an MT, they are applicable to any bilingual dictionary as long as one of the two languages is English or has a Wordnet linked to the Princeton Wordnet. Starting with 5 available bilingual dictionaries, we create 48 new bilingual dictionaries. Of these, 30 pairs of languages are not supported by the popular MTs: Google\footnote{http://translate.google.com/} and Bing\footnote{http://www.bing.com/translator}.
\end{quote}
\end{abstract}

\section{Introduction}
Bilingual dictionaries play a major role in applications such as machine translation, information retrieval, cross lingual document, automatic disambiguation of word sense, computing similarities among documents and increasing translation accuracy \cite{Knight1994}. Bilingual dictionaries are also useful to general readers, who may need help in translating documents in a given language to their native language or to a language in which they are familiar. Such dictionaries may also be important from an intelligence perspective, especially when they deal with smaller languages from sensitive areas of the world. Creating new bilingual dictionaries is also a purely intellectual and scholarly endeavor important to the humanities and other scholars. 

The powerful online MTs developed by Google and Bing provide pairwise translations for 80 and 50 languages, respectively. These machines provide translators for single words and phrases also. In spite of so much information for some ``privileged'' language pairs, there are many languages for which we are lucky to find a single bilingual dictionary online or in print. For example, we can find an online Karbi-English dictionary and an English-Vietnamese dictionary, but we can not find a Karbi-Vietnamese\footnote{The ISO 693-3 codes of Arabic, Assamese, Dimasa, English, Karbi and Vietnamese are \emph{arb}, \emph{asm}, \emph{dis}, \emph{eng}, \emph{ajz} and \emph{vie}, respectively.} dictionary.

The question we address in this paper is the following: Given a language, especially a resource-poor language, with only one available dictionary translating from that language to a resource-rich language, can we construct several good dictionaries translating from the original language to many other languages using publicly available resources such as bilingual dictionaries, MTs and Wordnets? We call a dictionary \emph{good} if each entry in it is of high quality and we have the largest number of entries possible. We must note that these two objectives conflict: Frequently if an algorithm produces a large number of entries, there is a high probability that the entries are of low quality.  Restating our goal, with only one input dictionary translating from a source language to a language which is a language with an available Wordnet linked to the Princeton Wordnet (PWN) \cite{Fellbaum1998}, we create a number of good bilingual dictionaries from that source language to  all other languages supported by an MT with different levels of accuracy and sophistication. 

Our contribution in this work is our reliance on the existence of just one bilingual dictionary between a low-resource language and a resource-rich language, viz., \emph{eng}. This strict constraint on the number of input bilingual dictionaries can be met by even many endangered languages. We consciously decided not to depend on additional bilingual dictionaries or external corpora because such languages usually do not have such resources. The simplicity of our algorithms along with low-resource requirements are our main strengths.

\section{Related work}

Let A, B and C be three distinct human languages. Given two input dictionaries \textit{Dict(A, B)} consisting of entries $(a_i, b_k)$ and \textit{Dict(B,C)} containing entries $(b_k,c_j)$, a na\"ive method to create a new bilingual dictionary \textit{Dict(A,C)} may use \emph{B} as a pivot: If a word $a_i$ is translated into a word $b_k$, and the word $b_k$ is translated into a word $c_j$, this straightforward approach concludes that the word $c_j$ is a translation of the word $a_i$, and adds the entry $(a_i,c_j)$ into the dictionary \textit{Dict(A,C)}. However, if the word $b_k$ has more than one sense, being a polysemous word, this method might make wrong conclusions. For example, if the word $b_k$ has two distinct senses which are translated into words $c_{j1}$ and $c_{j2}$, the straightforward method will conclude that the word $a_i$ is translated into both the word $c_{j1}$ and the word $c_{j2}$, which may be incorrect. This problem is called the ambiguous word sense problem. After obtaining an initial bilingual dictionary, past researchers have used several approaches to mitigate the effect of the ambiguity problem. All the methods used for word sense disambiguation use Wordnet distance between source and target words in some ways, in addition to looking at dictionary entries in forward and backward directions and computing the amount of overlap or match to obtain disambiguation scores \cite{Tanaka1994}, \cite{Gollins2001}, \cite{Bond2001}, \cite{Ahn2006}, \cite{Bond2008}, \cite{Mausam2010},  \cite{Lam2013} and \cite{Shaw2013}. The formulas used and the names used for the disambiguation scores by different authors are different. Researchers have also merged information from sources such as parallel corpora or comparable corpora  \cite{Nerima2008}, \cite{Otero2010} and a Wordnet \cite{Istvan2009}. Some researchers have also extracted bilingual dictionaries from parallel corpora or comparable corpora using statistical methods \cite{Brown1997}, \cite{Haghighi2008},   \cite{Nakov2009}, \cite{Heja2010}, \cite{Ljubesic2011} and \cite{Bouamor2013}.

The primary similarity among these methods is that they work with languages that already possess several lexical resources or these approaches take advantage of related languages (that have some lexical resources) by using such languages as intermediary. The accuracies of bilingual dictionaries created from available dictionaries and Wordnets are usually high. However, it is expensive to create such original lexical resources and they do not always exist for many languages. For example, such resources do not exist for most major languages of India, some spoken by hundred of millions. The same holds for many other widely spoken languages from around the world. In addition, these methods can only generate one or just a few new bilingual dictionaries using published approaches. 

In this paper, we propose methods for creating a significant number of bilingual dictionaries from a single available bilingual dictionary, which translates a source language to a resource-rich language with an available Wordnet. We use publicly available Wordnets in several resource-rich languages and a publicly available MT as well.

\section{Bilingual dictionary}

An entry in a dictionary, called \emph{LexicalEntry}, is a 2-tuple \emph{<LexicalUnit, Definition>}. A \emph{LexicalUnit} is a word or phrase being defined, the so-called \emph{definiendum} \cite{Landau1984}. A list of entries sorted by the \emph{LexicalUnit} is called a \emph{lexicon} or a \emph{dictionary}. Given a \emph{LexicalUnit}, the \emph{Definition} associated with it usually contains its class and pronunciation, its meaning, and possibly additional information. The meaning associated with it can have several \emph{Sense}s. A \emph{Sense} is a discrete representation of a single aspect of the meaning of a word. Entries in the dictionaries we create are of the form $<LexicalUnit, Sense_1>$, $<LexicalUnit, Sense_2>$,....

\section{Proposed approaches}
This section describes approaches to create new bilingual dictionaries \emph{Dict(S,D)}, each of which translates a word in language \emph{S} to a word or multiword expression in a destination language \emph{D}. Our starting point is just one existing bilingual dictionary \emph{Dict(S,R)}, where \emph{S} is the source language and \emph{R} is an ``intermediate helper'' language. We require that the language \emph{R} has an available Wordnet linked to the PWN. We do not think this is a big imposition since the PWN and other Wordnets are freely available for research purposes.

\subsection{Direct translation approach (DT)}
We first develop a direct translation method which we call the DT approach (see Algorithm \ref{alg:algorithm0}). The DT approach uses transitivity to create  new bilingual dictionaries from existing dictionaries and an MT. An existing dictionary \emph{Dict(S,R)} contains alphabetically sorted \emph{LexicalUnit}s in a source language \emph{S} and each has one or more \emph{Sense}s in the language \emph{R}. We call such a sense $Sense_{R}$. To create a new bilingual dictionary \emph{Dict(S,D)}, we simply take every pair \emph{<LexicalUnit,$Sense_{R}$>} in \emph{Dict(S,R)} and translate $Sense_{R}$ to \emph{D} to generate translation candidates \emph{candidateSet} (lines 2-4).  When there is no translation of $Sense_{R}$ in \emph{D}, we skip that pair \emph{<LexicalUnit,$Sense_{R}$>}. Each candidate in \emph{candidateSet} becomes a $Sense_D$ in language \emph{D} of that \emph{LexicalUnit}. We add the new tuple <$LexicalUnit,Sense_{D}$> to \emph{Dict(S,D)} (lines 5-7).  

\begin{algorithm}[!h]
\caption{DT algorithm}
Input: Dict(S, R)\\
Output: Dict(S, D) 
\begin{algorithmic}[1]
\STATE Dict(S, D) := $\phi$
\FORALL{$LexicalEntry$ $\in$ Dict(S, R)}
\FORALL{$Sense_R$ $\in$ LexicalEntry}
\STATE candiateSet= translate($Sense_R$,D)
\FORALL {candidate $\in$ candiateSet}
\STATE $Sense_{D_j}$ = candidate
\STATE  add tuple  \emph{<LexicalUnit,$Sense_{D_j}$>} to Dict(S,D)
\ENDFOR
\ENDFOR
\ENDFOR
\end{algorithmic}
 \label{alg:algorithm0}
\end{algorithm}

\begin{figure}[!h]
\centering
\includegraphics[width=0.51\textwidth]{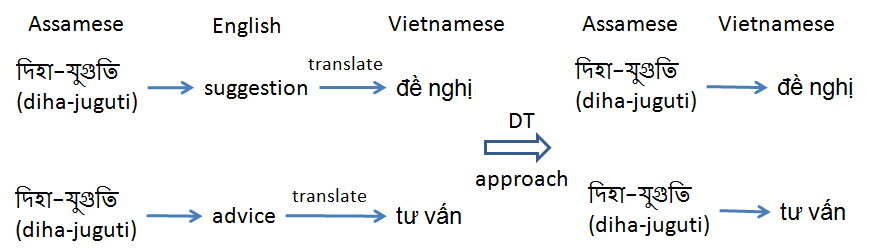}
\caption{\small{An example of DT approach for generating a new dictionary \emph{Dict(asm,eng)}. In \emph{Dict(asm,eng)}, the word ``diha-juguti'' in \emph{asm} has two translations in \emph{eng} ``suggestion'' and ``advice'', which are translated to \emph{vie} as ``{\fontencoding{T5}\selectfont
\dj \`\ecircumflex } {\fontencoding{T5}\selectfont ngh\d{i}}'' and   ``{\fontencoding{T5}\selectfont
t\uhorn } {\fontencoding{T5}\selectfont v\'\acircumflex n}'', respectively, using the Bing Translator. Therefore, in the new \emph{Dict(asm,vie)}, the word ``diha-juguti'' has two translations in \emph{vie} which are ``{\fontencoding{T5}\selectfont
\dj \`\ecircumflex } {\fontencoding{T5}\selectfont ngh\d{i}}'' and   ``{\fontencoding{T5}\selectfont
t\uhorn } {\fontencoding{T5}\selectfont v\'\acircumflex n}''.}}
\label{fig:exampledirect}
\end{figure}

An example of generating an entry for a \emph{Dict(asm,vie)} using the DT approach from an input \emph{Dict(asm,eng)} is presented in Figure \ref{fig:exampledirect}. 

\subsection{Using publicly available Wordnets as intermediate resources (IW)}
\begin{figure}[!h]
\centering
\includegraphics[width=0.5\textwidth]{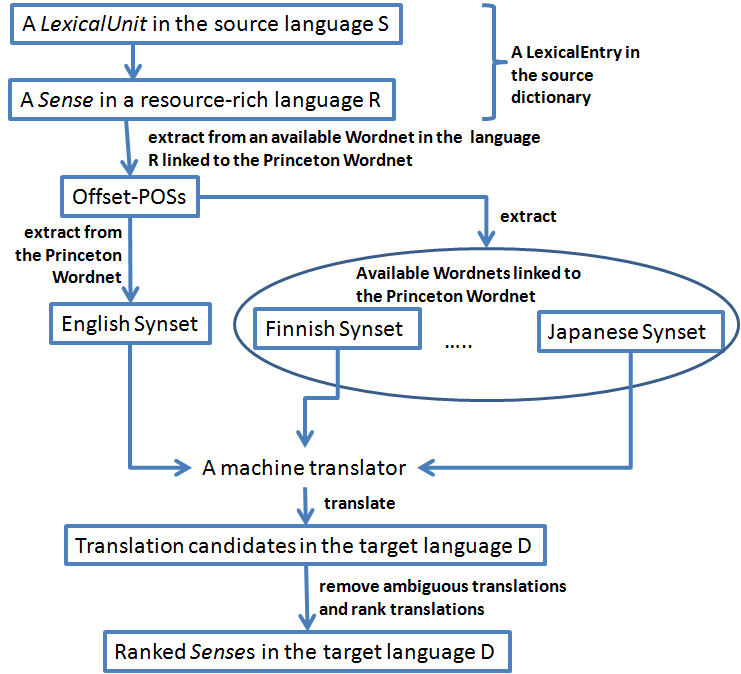} 
\caption{\small{The IW approach for creating a new bilingual dictionary}}
\label{fig:algorithmCreatingNewDicsUsingWN}
\end{figure}

To handle ambiguities in the dictionaries created, we propose the IW approach as in  Figure \ref{fig:algorithmCreatingNewDicsUsingWN} and Algorithm \ref{alg:alg1}. 

For each $Sense_R$ in every given \emph{LexicalEntry} from \emph{Dict(S,R)}, we find all \emph{Offset-POS}s\footnote{Synset is a set of cognitive synonyms. \emph{Offset-POS} refers to the offset for a synset with a particular POS, from the beginning of its data file. Words in a synset have the same sense.} in the  Wordnet of the language \textit{R} to which $Sense_{R}$ belongs (Algorithm \ref{alg:alg1}, lines 2-5). Then, we find a candidate set for translations from the \emph{Offset-POS}s and the destination language $D$ using Algorithm \ref{alg:alg2}. For each $\emph{Offset-POS}$ from the extracted $\emph{Offset-POS}s$, we obtain each $word$ belonging to that $\emph{Offset-POS}$ from different Wordnets  (Algorithm \ref{alg:alg2}, lines 2-3) and translate it to \emph{D} using a MT to generate translation candidates (Algorithm \ref{alg:alg2}, line 4). We add translation candidates to the $candidateSet$ (Algorithm \ref{alg:alg2}, line 6). Each candidate in the $candidateSet$ has 2 attributes: a translation of the word $word$ in the target language \emph{D}, the so-called $candidate.word$ and the occurrence count or the rank value of the $candidate.word$, the so-called $candidate.rank$. A candidate with a greater rank value is more likely to become a correct translation. Candidates having the same ranks are treated similarly. Then, we sort all candidates in the $candidateSet$ in descending order based on their rank values (Algorithm \ref{alg:alg1}, line 7), and add them into the new dictionary \emph{Dict(S,D)} (Algorithm \ref{alg:alg1}, lines 8-10). We can vary the Wordnets and the numbers of Wordnets used during experiments, producing different results.

\begin{algorithm}[!h]
\caption{IW algorithm}
Input: Dict(S,R)\\
Output: Dict(S, D)
\begin{algorithmic}[1]
\STATE Dict(S, D) := $\phi$
\FORALL{LexicalEntry $\in$ Dict(S,R)}
\FORALL{$Sense_R \in$ LexicalEntry}
\STATE candidateSet := $\phi$
\STATE Find all \emph{Offset-POS}s of synsets containing $Sense_R$ from the \emph{R} Wordnet
\STATE $candidatSet$ = FindCandidateSet (\emph{Offset-POSs}, D)
\STATE sort all $candidate$ in descending order based on their rank values
\FORALL {$candidate \in candidateSet$}

\STATE $Sense_D$=candidate.word
\STATE  add tuple \emph{<LexicalUnit,$Sense_D$>} to Dict(S,D)

\ENDFOR
\ENDFOR
\ENDFOR
\end{algorithmic}
 \label{alg:alg1}
\end{algorithm}

\begin{algorithm}[!h]
\caption{FindCandidateSet (\emph{Offset-POS}s,D)}
Input: \emph{Offset-POS}s, D\\
Output: candidateSet
\begin{algorithmic}[1]
\STATE $candidateSet$ := $\phi$
\FORALL{$\emph{Offset-POS} \in \emph{Offset-POS}s$}
\FORALL {$word$ in the $\emph{Offset-POS}$ extracted from the PWN and other available Wordnets linked to the PWN}
\STATE $candidate.word$= translate $(word,D)$
\STATE $candidate.rank$++
\STATE $candidateSet$ += $candidate$
\ENDFOR
\ENDFOR
\STATE return $candidateSet$
\end{algorithmic}
\label{alg:alg2}
\end{algorithm}

\begin{figure}[!h]
\centering
\includegraphics[width=0.5\textwidth]{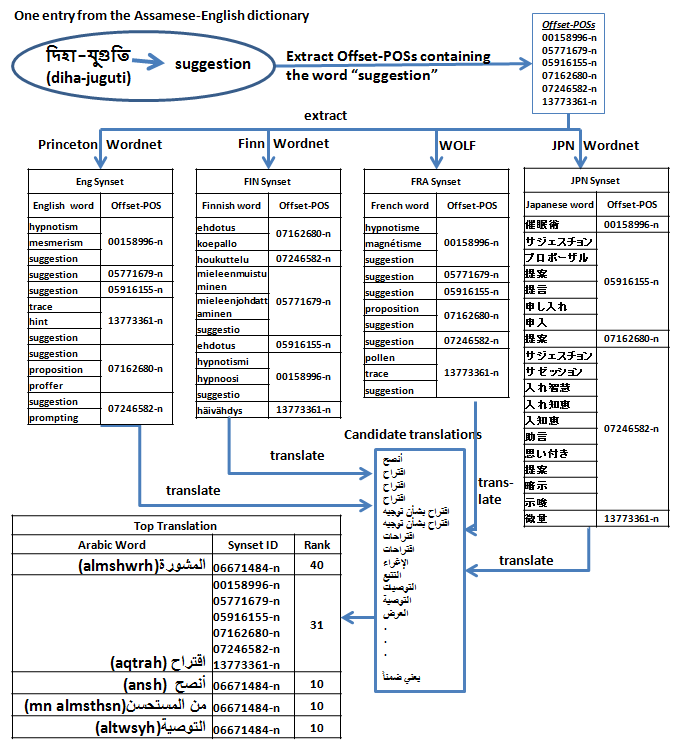}
\caption{\small{Example of generating new \emph{LexicalEntry}s for \emph{Dict(asm,arb)} using the IW approach from \emph{Dict(asm,eng)}. The word "diha-juguti" in \emph{asm} has two senses as in Figure \ref{fig:algorithmCreatingNewDicsUsingWN}: ``suggestion'' and `` advice''. This example only shows the IW approach to find the translation of "diha-juguti" with the sense ``suggestion''. We find all \emph{offset-POS}s in the PWN containing "suggestion". Then, we extract words belonging to all \emph{offset-POS}s from the PWN, FinnWordnet, WOLF and Japanese Wordnet. Next, we translate extracted words to \emph{arb} and rank them based on the occurrence counts. According to the ranks, the best translation of  "diha-juguti" in \emph{asm}, which has the greatest rank value, is the word "almshwrh" in \emph{arb}.}}
\label{fig:example}
\end{figure}

Figure \ref{fig:example} shows an example of creating entries for \emph{Dict(asm,arb)} from \emph{Dict(asm,eng)} using the IW approach.

\section{Experimental results}
\subsection{Data sets used}

Our approach is general, but to demonstrate the effectiveness and usefulness of our algorithms, we have carefully selected a few languages for experimentation. These languages include widely-spoken languages with limited computational resources such as \emph{arb} and \emph{vie}; a language spoken by tens of millions in a specific region within India, viz., \emph{asm} with almost no resources; and a couple of languages in the UNESCO's list of endangered languages, viz., \emph{dis} and \emph{ajz} both from northeast India, again with almost no resources at all.

We work with 5 existing bilingual dictionaries that translate a given language to a resource-rich language, which happens to be \emph{eng} in our experiments: \emph{Dict(arb,eng)} and \emph{Dict(vie,eng)} supported by Panlex\footnote{http://panlex.org/}; \emph{Dict(ajz,eng)} and \emph{Dict(dis,eng)} supported by Xobdo\footnote{http://www.xobdo.org/}; one \emph{Dict(asm,eng)} created by integrating two dictionaries \emph{Dict(asm,eng)} provided by Panlex and Xobdo. The numbers of entries in \emph{Dict(ajz,eng)}, \emph{Dict(arb,eng)}, \emph{Dict(asm-eng)}, \emph{Dict(dis,eng)} and \emph{Dict(vie,eng)} are 4682,  53194, 76634, 6628 and 231665, respectively. Many \emph{LexicalEntry}s in some of our input dictionaries have no POS information. For example, 100\% and 6.63\% \emph{LexicalEntry}s in \emph{Dict(arb,eng)} and \emph{Dict(vie,eng)}, respectively, do not have POS.

To solve the problem of ambiguities, we use the PWN and Wordnets in several other languages linked to the PWN provided by the Open Multilingual Wordnet project \cite{Bond2013}: FinnWordnet \cite{Linden2010} (FWN), WOLF \cite{Sagot2008} (WWN) and Japanese Wordnet \cite{Isahara2008} (JWN). We choose these Wordnets because they are already aligned with the PWN, cover a large number of synsets of the about 5,000 most frequently used word senses and are available online\footnote{http://compling.hss.ntu.edu.sg/omw/} for free. Depending on which Wordnets are used and the number of intermediate Wordnets, the sizes and qualities of the new dictionaries created change. The Microsoft Translator Java API\footnote{https://datamarket.azure.com/dataset/bing/microsofttranslator} is used as another main resource. The Microsoft Translator supports translations for 50 languages.

\begin{figure}[h]
\centering
\includegraphics[width=0.5\textwidth]{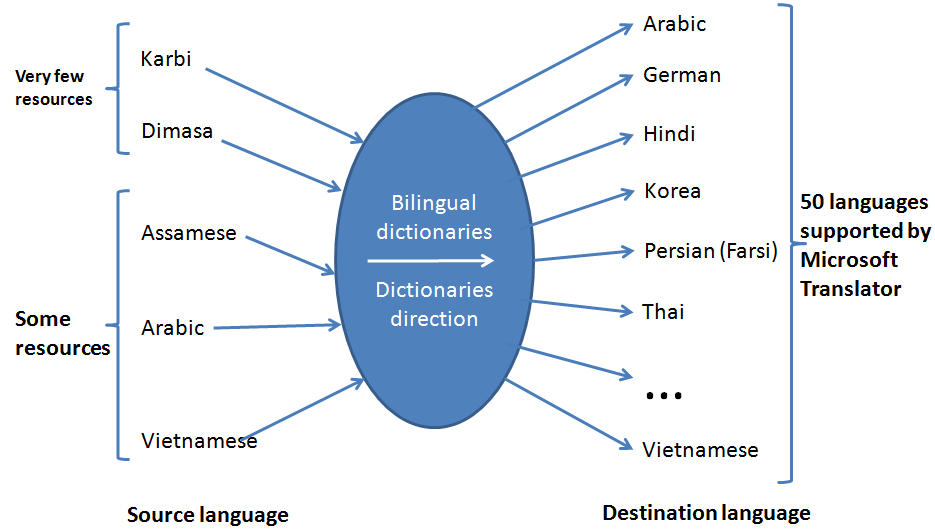}
\caption{New bilingual dictionaries created}
\label{fig:creatingDictionaries}
\end{figure}

In our experiments, we create dictionaries from any of \{\emph{ajz, arb, asm, dis, vie}\} to any non-\emph{eng} language supported by the Microsoft Translator, e.g., \emph{arb}, Chinese (\emph{cht}), German (\emph{deu}), Hmong Daw (\emph{mww}), Indonesian (\emph{ind}), Korean (\emph{kor}), Malay (\emph{zlm}), Thai (\emph{tha}), Spanish (\emph{spa}) and \emph{vie}, as shown in  Figure \ref{fig:creatingDictionaries}.

 \subsection{Results and human evaluation} 

Ideally, evaluation should be performed by volunteers who are fluent in both source and target languages. However, for evaluating created dictionaries, we could not recruit any individuals who are experts in two appropriate languages\footnote{This is not surprising. Considering languages we focus on are disparate belonging to different classes, with provenance spread out around the world, and frequently resource poor and even endangered. For example, it is almost impossible to find an individual fluent in the endangered language Karbi and Vietnamese.}. Hence, every dictionary is evaluated by 2 people sitting together, one using the target language as mother tongue, and the other the source language. Each volunteer pair was requested to evaluate using a 5-point scale -- 5: excellent, 4: good, 3: average, 2: fair and 1: bad. For selecting the samples to evaluate, we follow the concept of ``simple random method'' \cite{Hays1971}. According to the general rules of thumb, we can be confident of the normal approximation whenever the sample size is at least 30 \cite[page 310]{Ross2010}. To achieve reliable judgment, we randomly picked 100 translations from every dictionary created.

To study the effect of available resources used to create our new dictionaries, we first evaluate the input dictionaries. The average score of \emph{LexicalEntry}s in the input dictionaries \emph{Dict(arb,eng)}, \emph{Dict(asm,eng)} and \emph{Dict(vie,eng)} are 3.58, 4.65 and 3.77, respectively. This essentially means that \emph{Dict(asm,eng)} is of almost excellent quality, while the other two are of reasonably good quality. These are the best dictionaries we could find. The average score and the number of \emph{LexicalEntry}s in the dictionaries we create using the DT approach are presented in Table \ref{tab:directDic}. We believe that if we find ``better'' input dictionaries, our results will be commensurately better.

\begin{table}[!h]
 \centering
\begin{tabular}{|l|c|r|c|c|r|}
\hline    
Dict.&Score&Entries&Dict.&Score&Entries\\ \hline 
arb-deu&4.29&1,323&arb-spa&3.61&1,709\\  
arb-vie&3.66&2,048&asm-arb&4.18&47,416\\  
asm-spa&4.81&20,678&asm-vie&4.57&42,743\\  
vie-arb&2.67&85,173&vie-spa&3.55&35,004\\ \hline

\end{tabular}
\caption{The average score and the number of \emph{LexicalEntry}s in the dictionaries created using the DT approach.}
\label{tab:directDic}
\end{table}

The average scores and the numbers of \emph{LexicalEntry}s in the dictionaries created by the IW approach are presented in Table \ref{tab:AvgScore} and Table \ref{tab:NumEntries}, respectively. In these tables, \emph{Top n }means dictionaries created by picking only translations with the top\emph{ n} highest ranks for each word, \emph{A}: dictionaries created using PWN only; \emph{B}: using PWN and FWN; \emph{C}: using PWN, FWN and JWN; \emph{D}: using PWN, FWN, JWN and WWN.  The method using all 4 Wordnets produces dictionaries with the highest scores and the highest number of \emph{LexicalEntry}s as well.

\begin{table}[!h]
 \centering
\begin{tabular}{|c|c|c|c|c|c|}
\hline    
\multicolumn{2}{|c|}{Dict.}& \multicolumn{4}{c|}{Wordnets used} \\ \cline{3-6}
\multicolumn{2}{|c|}{we create}&A & B & C & D\\ \hline
\multirow{3}{*}{arb-vie}&Top 1 &3.42&3.65&3.33&\textbf{3.71}\\  \cline{2-6}
&Top 3 &3.33&3.58&\textbf{3.76}&3.61\\   \cline{2-6}
&Top 5 &2.99&3.04&3.08&\textbf{3.31}\\ \hline 

\multirow{3}{*}{asm-arb}&Top 1 &4.51&3.83&\textbf{4.69}&4.67\\  \cline{2-6}
&Top 3 &4.03&3.75&3.80&\textbf{4.10}\\   \cline{2-6}
&Top 5 &3.78&3.85&3.42&\textbf{4.00}\\ \hline

\multirow{3}{*}{asm-vie}&Top 1 &4.43&4.31&3.86&\textbf{4.43}\\  \cline{2-6}
&Top 3 &3.93&3.59&3.33&\textbf{3.94}\\  \cline{2-6}
&Top 5 &\textbf{3.74}&3.34&3.4&2.91\\ \hline

\multirow{3}{*}{vie-arb}&Top 1 &3.11&2.94&2.78&\textbf{3.11}\\  \cline{2-6}
&Top 3 &2.47&2.72&2.61&\textbf{3.02}\\   \cline{2-6}
&Top 5 &2.54&2.37&2.60&\textbf{2.73}\\ \hline
\end{tabular}\vspace{-2mm}
\caption{The average score of \emph{LexicalEntry}s in the dictionaries we create using the IW approach.}
\label{tab:AvgScore}
\end{table}

\begin{table}[!h]
 \centering
\begin{tabular}{|>{\small}c|>{\small}c|>{\small}r|>{\small}r|>{\small}r|>{\small}r|}
\hline    
 
\multicolumn{2}{|c|}{Dict.}& \multicolumn{4}{c|}{Wordnets used} \\ \cline{3-6}
\multicolumn{2}{|c|}{we create}&A & B & C & D\\ \hline
arb&Top1 &1,786&2,132&2,169&\textbf{2,200}\\  \cline{2-6}
-&Top3 &3,434&4,611&4,908&\textbf{5,110}\\   \cline{2-6}
 vie&Top5 &4,123&5,926&6,529&\textbf{6,853}\\ \hline 

asm&Top1 &27,039&27,336&27,449&\textbf{27,468}\\  \cline{2-6}
-&Top3 &70,940&76,695&78,979&\textbf{79,585}\\   \cline{2-6}
 arb&Top5 &104,732&118,261&125,087&\textbf{126,779}\\ \hline

asm&Top1 &25,824&26,898&27,064&\textbf{27,129}\\  \cline{2-6}
-&Top3 &64,636&73,652&76,496&\textbf{77,341}\\  
 vie&Top5 &92,863&111,977&120,090&\textbf{122,028}\\ \hline

vie&Top1 &63,792&65,606&\textbf{66,040}&65,862\\  \cline{2-6}
-&Top3 &152,725&177,666&183,098&\textbf{185,221}\\   \cline{2-6}
 arb&Top5 &210,220&261,392&278,117&\textbf{282,398}\\ \hline
\end{tabular}
\caption{The number of \emph{LexicalEntry}s in the dictionaries we create using the IW approach.}\vspace{-2mm}
\label{tab:NumEntries}
\end{table}

The number of \emph{LexicalEntry}s and the accuracies of the newly created dictionaries definitely depend on the sizes and qualities of the input dictionaries. Therefore, if the sizes and the accuracies of the dictionaries we create are comparable to those of the input dictionaries, we conclude that the new dictionaries are acceptable. Using four Wordnets as intermediate resources to create new bilingual dictionaries increases not only the accuracies but also the number of \emph{LexicalEntry}s in the dictionaries created. We also evaluate several bilingual dictionaries we create for a few of the language pairs. Table \ref{tab:Dic2} presents the number of \emph{LexicalEntry}s and the average score of some of the bilingual dictionaries generated using the four Wordnets. 

\begin{table}[!h]
 \centering
\begin{tabular}{|l|c|r|c|r|}
\hline    
Dict.&\multicolumn{2}{c|}{Top 1}&\multicolumn{2}{c|}{Top 3}\\  \cline{2-5}
we create& Score & Entries& Score & Entries \\ \hline
arb-deu&4.27&1,717&4.21 &3,859\\ 
arb-spa&4.54& 2,111&  4.27& 4,673\\  
asm-spa&4.65&26,224&4.40  & 72,846\\  
vie-spa&3.42&61,477&3.38 &159,567\\ \hline
\end{tabular}
\caption{The average score of entries and the number of \emph{LexicalEntry}s in some other bilingual dictionaries constructed using 4 Wordnets: PWN, FWN, JWN and WWN.}
\label{tab:Dic2}
\end{table}

The dictionaries created from \emph{arb} to other languages have low accuracies because our algorithms rely on the POS of the \emph{LexicalUnit}s to find the \emph{Offset-POS}s and the input \emph{Dict(arb,eng)} does not have POS. We were unable to access a better \emph{Dict(arb,eng)} for free. For \emph{LexicalEntry}s without POS, our algorithms choose the best POS of the \emph{eng} word. For instance, the word ``book'' has two POSs, viz., ``verb'' and ``noun'', of which ``noun'' is more common. Hence, all translations to the word ``book'' in \emph{Dict(arb,eng)} will have the same POS ``noun''. As a result, all \emph{LexicalEntry}s translating to the word ``book'' will be treated as a noun, leading to many wrong translations.

Based on experiments, we conclude that using the four public Wordnets, viz., PWN, FWN, JWN and WWN as intermediate resources, we are able to create good bilingual dictionaries, considering the dual objective of high quality and a large number of entries. In other words, the IW approach using the four intermediate Wordnets is our best approach. We note that if we include only translations with the highest ranks, the resulting dictionaries have accuracies even better than the input dictionaries used. We are in the processes of finding volunteers to evaluate dictionaries translating from \emph{ajz} and \emph{dis} to other languages. Table \ref{tab:Dic3} presents the number of entries of some of dictionaries, we created using the best approach, without human evaluation.

\begin{table}[!h]
 \centering
\begin{tabular}{|l|r|l|r|}
\hline    
Dict.&Entries&Dict.&Entries\\ \hline
ajz-arb&4,345&ajz-cht&3,577\\
ajz-deu&3,856&ajz-mww&4,314\\
ajz-ind&4,086&ajz-kor&4,312\\ 
ajz-zlm&4,312&ajz-spa&3,923\\
ajz-tha&4,265&ajz-vie &4,344\\
asm-cht &67,544 &asm-deu&71,789\\
asm-mww&79,381&asm-ind&71,512\\
asm-kor&79,926&asm-zlm&80,101\\
asm-tha&78,317&dis-arb& 7,651\\
dis-cht&6,120&dis-deu&6,744\\
dis-mww&7,552&dis-ind&6,762\\
dis-kor&7,539& dis-zlm&7,606\\
dis-spa&6,817& dis-tha&7,348\\
dis-vie&7,652&&\\\hline
\end{tabular}
\caption{The number of \emph{LexicalEntry}s in some other the dictionaries, we created using the best approach. \textit{ajz} and \textit{dis} are endangered.
\label{tab:Dic3}}
\end{table}

\subsection{Comparing with existing approaches}
It is difficult to compare approaches because the language involved in different papers are different, the number and quality of input resources vary and the evaluation methods are not standard. However, for the sake of completeness, we make an attempt at comparing our results with \cite{Istvan2009}. The precision of the best dictionary created by  \cite{Istvan2009} is 79.15\%. Although our score is not in terms of percentage, we obtain the average score of all dictionaries we created using 4 Wordnets and containing 3-top greatest ranks \emph{LexicalEntry}s is 3.87/5.00, with the highest score being 4.10/5.00 which means the entries are very good on average. If we look at the greatest ranks only (Top 1 ranks), the highest score is 4.69/5.00 which is almost excellent. We believe that we can apply these algorithms to create dictionaries where the source is any language, with a bilingual dictionary, to \textit{eng}.

To handle ambiguities, the existing methods need at least two intermediate dictionaries translating from the source language to intermediate languages. For example, to create \emph{Dict(asm,arb)}, \cite{Gollins2001} and \cite{Mausam2010} need at least two dictionaries: \emph{Dict(asm,eng)} and \emph{Dict(asm, French)}. For \emph{asm}, the second dictionary simply does not exist to the best of our knowledge. The IW approach requires only one input dictionary. This is a strength of our method, in the context of resources-poor language.

\subsection{Comparing with Google Translator}
Our purpose of creating dictionaries is to use them for machine learning and machine translation. Therefore, we evaluate the dictionaries we create against a well-known high quality MT: the Google Translator. We do not compare our work against the Microsoft Translator because we use it as an input resource.

We randomly pick 300 \emph{LexicalEntry}s from each of our created dictionaries for language pairs supported by the Google Translator. Then, we compute the matching percentages between translations in our dictionaries and translations from the Google Translator. For example, in our created dictionary \emph{Dict(vie,spa)}, ``{\fontencoding{T5}\selectfont
 ng\uhorn \`\ohorn i qu\h{a}ng} c\'ao'' in \emph{vie} translates to  ``anunciante'' in \emph{spa} which is as same as the translation given by the Google Translator. As the result, we mark the \emph{LexicalEntry} (``{\fontencoding{T5}\selectfont
 ng\uhorn \`\ohorn i qu\h{a}ng} c\'ao'', ``anunciante'') as ``matching''. The matching percentages of our dictionaries \emph{Dict(arb,spa)}, \emph{Dict(arb,vie)}, \emph{Dict(arb,deu)}, \emph{Dict(vie,deu)}, \emph{Dict(vie,spa)} and the Google Translator are 55.56\%, 39.16\%, 58.17\%, 25.71\%, and 35.71\%, respectively.

The \emph{LexicalEntry}s  marked as ``unmatched'' do not mean our translations are incorrect. Table \ref{tab:UnMatchGoogle} presents some \emph{LexicalEntry}s which are correct but are marked as ``unmatched''.

\begin{table}[!h]
 \centering
\begin{tabular}{c}
\includegraphics[width=0.47\textwidth]{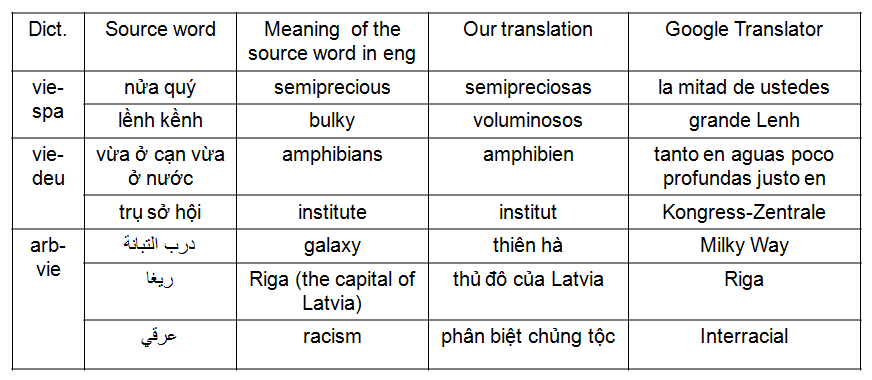}
\end{tabular}
\caption{Some \emph{LexicalEntry}s in dictionaries we created are correct but do not match with translations from the Google Translator. According to our evaluators, the translations from the Google Translator of the first four source words are bad.}
\label{tab:UnMatchGoogle}
\end{table}

\vspace{-5mm}
\subsection{Crowdsourcing for evaluation}
To achieve better evaluation, we intend to use crowdsourcing for evaluation. We are in the process of creating a Website where all dictionaries we create will be available, along with a user friendly interface to give feedback on individual entries. Our goal will be to use this feedback to improve the quality of the dictionaries.

\section{Conclusion} 
We present two approaches to create a large number of \emph{good} bilingual dictionaries from only one input dictionary, publicly available Wordnets and a machine translator. In particular, we created 48 new bilingual dictionaries from 5 input bilingual dictionaries. We note that 30 dictionaries we created have not supported by any machine translation yet. We believe that our research will help increase significantly the number of resources for machine translators which do not have many existing resources or are not supported by machine translators. This includes languages such as \emph{ajz}, \emph{asm} and \emph{dis}, and tens of similar languages. We use Wordnets as intermediate resources to create new bilingual dictionaries because these Wordnets are available online for unfettered use and they contain information that can be used to remove ambiguities. 

\section{Acknowledgments}
We would like to thank the volunteers evaluating the dictionaries we create: Dubari Borah, Francisco Torres Reyes, Conner Clark, and Tri Si Doan. We also thank all friends in the Microsoft, Xobdo and Panlex projects who provided us dictionaries.

\bibliography{Bibliography}
\bibliographystyle{aaai}
\end{document}